\documentclass{article}

\usepackage{amssymb}
\usepackage{graphics} 
\usepackage{graphicx}
\usepackage{amsmath}
\usepackage{booktabs}
\usepackage{multirow}
\usepackage{booktabs,caption}
\usepackage[flushleft]{threeparttable}
\usepackage{multicol}
\usepackage{array}

\newcolumntype{M}[1]{>{\centering\arraybackslash}m{#1}}
\newcolumntype{N}{@{}m{0pt}@{}}

\usepackage[final]{corl_2020} % Uncomment for the camera-ready ``final'' version.

\title{TartanVO: A Generalizable Learning-based VO}
\author{
  Wenshan Wang\thanks{Corresponding author: \url{wenshanw@andrew.cmu.edu}} \\
  Carnegie Mellon University\\
  \And
  Yaoyu Hu\\
  Carnegie Mellon University\\
  \And
  Sebastian Scherer\\
  Carnegie Mellon University\\
}

\begin{document}
\maketitle

%===============================================================================
% \vspace{-0.5in}
\begin{abstract}
    We present the first learning-based visual odometry (VO) model, which generalizes to multiple datasets and real-world scenarios, and outperforms geometry-based methods in challenging scenes. 
    We achieve this by leveraging the SLAM dataset TartanAir, which provides a large amount of diverse synthetic data in challenging environments. 
    Furthermore, to make our VO model generalize across datasets, we propose an up-to-scale loss function and incorporate the camera intrinsic parameters into the model. 
    Experiments show that a single model, TartanVO, trained only on synthetic data, \textit{without any finetuning}, can be generalized to real-world datasets such as KITTI and EuRoC, demonstrating significant advantages over the geometry-based methods on challenging trajectories.
    Our code is available at \url{https://github.com/castacks/tartanvo}.  
    %  achieve the state-of-the-art results with one single model  on 
\end{abstract}


%===============================================================================

\section{Introduction}
	
Visual SLAM (Simultaneous Localization and Mapping) becomes more and more important for autonomous robotic systems due to its ubiquitous availability and the information richness of images~\citep{fuentes2015visual}.
Visual odometry (VO) is one of the fundamental components in a visual SLAM system.
Impressive progress has been made in both geometric-based methods~\citep{Engel2014lsd, mur2015orb,forster2014svo,engel2017direct} % klein2007parallel,engel2013semi,
and learning-based methods~\citep{Zhou2017SfMLearner, Vija2017SfMNet, wang2018end, wang2019improving}. % pillai2017towards, costante2018ls, iyer2018geometric,
However, it remains a challenging problem to develop a robust and reliable VO method for real-world applications.

% Real-life environments are full of inconsistencies such as changing light, low illumination, dynamic objects, and texture-less scenes. 

On one hand, geometric-based methods are not robust enough in many real-life situations 
% such as varying or low illumination, dynamic objects, and texture-less scenes
\citep{younes2017keyframe,tartanair2020iros}. On the other hand, although learning-based methods demonstrate robust performance on many visual tasks, including object recognition, semantic segmentation, depth reconstruction, and optical flow, we have not yet seen the same story happening to VO. 
% #ne of the reasons may be the generalization ability of learning-based models.  

It is widely accepted that by leveraging a large amount of data, deep-neural-network-based methods can learn a better feature extractor than engineered ones, resulting in a more capable and robust model. But why haven't we seen the deep learning models outperform geometry-based methods yet? We argue that there are two main reasons. First, \textit{the existing VO models are trained with insufficient diversity}, which is critical for learning-based methods to be able to generalize. By diversity, we mean diversity both in the scenes and motion patterns. For example, a VO model trained only on outdoor scenes is unlikely to be able to generalize to an indoor environment. Similarly, a model trained with data collected by a camera fixed on a ground robot, with limited pitch and roll motion, will unlikely be applicable to drones. Second, most of the current learning-based VO models \textit{neglect some fundamental nature of the problem}
% which are well understood by computer vision and robotics communities
\textit{which is well formulated in geometry-based VO theories}. From the theory of multi-view geometry, we know that recovering the camera pose from a sequence of monocular images has \textit{scale ambiguity}. Besides, recovering the pose needs to take account of the camera intrinsic parameters (referred to as the \textit{intrinsics ambiguity} later). Without explicitly dealing with the scale problem and the camera intrinsics, a model learned from one dataset would likely fail in another dataset, no matter how good the feature extractor is. 

% 1. Lack of training data
% 2. Disrespect the nature of VO problem: try to recover scale from the monocular image
% 3. Inconsistency of camera intrinsic values

To this end, we propose a learning-based method that can solve the above two problems and can generalize across datasets. Our contributions come in three folds.
First, we demonstrate the crucial effects of data diversity on the generalization ability of a VO model by comparing performance on different quantities of training data.
Second, we design an up-to-scale loss function to deal with the scale ambiguity of monocular VO.
Third, we create an intrinsics layer (IL) in our VO model enabling generalization across different cameras.
% The experiments show the effectiveness of our design. 
To our knowledge, our model is the first learning-based VO that has competitive performance in various real-world datasets without finetuning. Furthermore, compared to geometry-based methods, our model is significantly more robust in challenging scenes. A demo video can be found: \url{https://www.youtube.com/watch?v=NQ1UEh3thbU}

% First, we utilize the recently released sythetic dataset TartanAir~\citep{tartanair2020iros}, which provides large scale data with diverse realistic scenes and motion patterns. Second, we design an up-to-scale loss function, and an intrinsics layer (IL), which enable our model to work with image sequences collected by different cameras. In our experiments, we demonstrate the crucial effects of the diverse training data and those new designs. We also show for the first time, the learning-based model outperform the geometry-based methods on multiple real-world datasets, especially in challenging scenes. 

% 1. Utilizing large amount of training data 
% 2. Incorporating ideas from geometric-based methods: Modular network structure, up-to-scale loss.
% 3. Incorporating camera intrinsic values

%===============================================================================

\section{Related Work}
\label{sec:citations}
\vspace{-0.1in}

Besides early studies of learning-based VO models \citep{roberts2008memory, guizilini2012semi, ciarfuglia2014evaluation, tateno2017cnn}, more and more end-to-end learning-based VO models have been studied with improved accuracy and robustness. The majority of the recent end-to-end models adopt the unsupervised-learning design \citep{Zhou2017SfMLearner, yin2018geonet, zhan2018unsupervised, Ranjan_2019_CVPR}, due to the complexity and the high-cost associated with collecting ground-truth data. However, supervised models trained on labeled odometry data still have a better performance \citep{costante2016exploring, Ummenhofer_2017_CVPR}.

To improve the performance, end-to-end VO models tend to have auxiliary outputs related to camera motions, such as depth and optical flow. 
% Depth prediction is particularly useful with its 3D geometric nature and is widely used among current unsupervised VO models. 
With depth prediction, models obtain supervision signals by imposing depth consistency between temporally consecutive images \citep{zhan2018unsupervised, Yang_2020_CVPR}. This procedure can be interpreted as matching the temporal observations in the 3D space. A similar effect of temporal matching can be achieved by producing the optical flow, e.g., \cite{yin2018geonet, Zou_2018_ECCV, Ranjan_2019_CVPR} jointly predict depth, optical flow, and camera motion. 

Optical flow can also be treated as an intermediate representation that explicitly expresses the 2D matching. Then, camera motion estimators can process the optical flow data rather than directly working on raw images\cite{Ummenhofer_2017_CVPR, Zhou_2018_ECCV}. If designed this way, components for estimating the camera motion can even be trained separately on available optical flow data \cite{costante2016exploring}. 
% The intermediate 2D representation can also stem from sparse features \cite{detone2018self, tang2019gcnv2}. Some models choose to not have any interpretable intermediate representation and directly make motion predictions \cite{wang2017deepvo}. 
We follow these designs and use the optical flow as an intermediate representation.

It is well known that monocular VO systems have scale ambiguity. Nevertheless, most of the supervised learning models did not handle this issue and directly use the difference between the model prediction and the true camera motion as the supervision \cite{Ummenhofer_2017_CVPR, tang2018ba, clark2018ls}. In \cite{costante2016exploring}, the scale is handled by dividing the optical flow into sub-regions and imposing a consistency of the motion predictions among these regions. 
% Some work utilize stereo input to deal with real-world scale ambiguity \cite{zhan2018unsupervised,Wang_2019_CVPR}. 
In non-learning methods, scale ambiguity can be solved if a 3D map is available \cite{li_2020_ICRA}. 
% Depth prediction helps on correcting scale-drifts.
\citet{Ummenhofer_2017_CVPR} introduce the depth prediction to correcting the scale-drift. \citet{Tateno_2017_CVPR} and \citet{ Sheng_2019_ICCV} ameliorate the scale problem by leveraging the key-frame selection technique from SLAM systems. Recently, \citet{zhan_2020_ICRA} use PnP techniques to explicitly solve for the scale factor.
% , however, the camera motion is recovered by non-learning method. 
The above methods introduce extra complexity to the VO system, however, the scale ambiguity is not totally suppressed for monocular setups especially in the evaluation stage. 
Instead, some models choose to only produce up-to-scale predictions.
% and try to find new ways to make the system more robust to scale issues. 
\citet{Wang_2018_CVPR} reduce the scale ambiguity in the monocular depth estimation task by normalizing the depth prediction before computing the loss function. Similarly, we will focus on predicting the translation direction rather than recovering the full scale from monocular images, by defining a new up-to-scale loss function. 

\begin{figure} [b]
	\begin{center}
	\vspace{-0.1in}
		\includegraphics[width=1.0\textwidth]{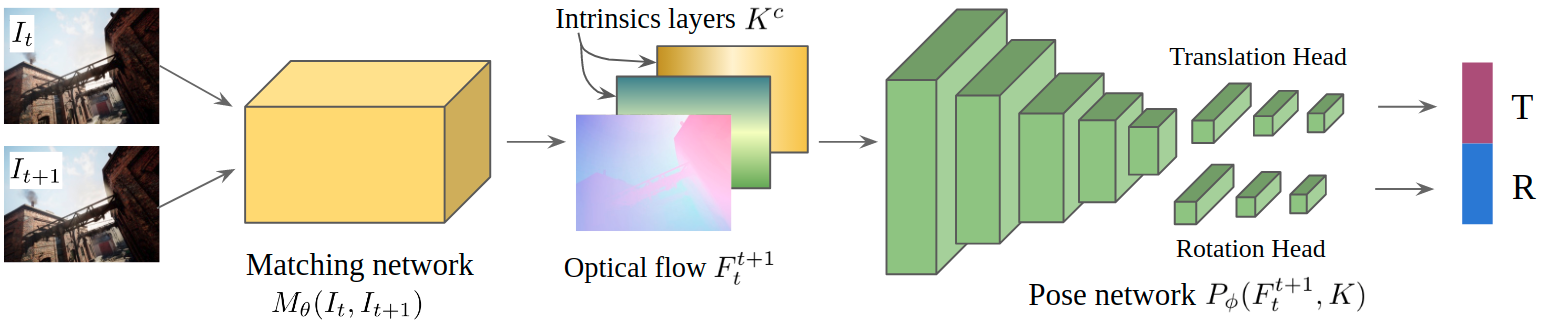}
	\end{center}
	\vspace{-0.1in}
	\caption{The two-stage network architecture. The model consists of a matching network, which estimates optical flow from two consecutive RGB images, followed by a pose network predicting camera motion from the optical flow.}
	\label{Fig:arch}
	\vspace{-0.2in}
\end{figure}

Learning-based models suffer from generalization issues when tested on images from a new environment or a new camera. Most of the VO models are trained and tested on the same dataset \cite{yin2018geonet, zhan2018unsupervised, Wang_2019_CVPR, Ranjan_2019_CVPR}. Some multi-task models \cite{Zhou2017SfMLearner, Ummenhofer_2017_CVPR, Mahjourian_2018_CVPR, Zou_2018_ECCV} only test their generalization ability on the depth prediction, not on the camera pose estimation. Recent efforts, such as \cite{Li_2020_CVPR}, use model adaptation to deal with new environments, however, additional training is needed on a per-environment or per-camera basis. In this work, we propose a novel approach to achieve cross-camera/dataset generalization, by incorporating the camera intrinsics directly into the model. 
% To the authors' knowledge, the most similar work is \cite{Tateno_2017_CVPR} where depth predictions are normalized by referring to the intrinsics of the underlying camera. In our model, we enable the cross-camera generalization capability by encoding the camera intrinsics directly.

%===============================================================================

\section{Approach}
\label{sec:approach}

\subsection{Background} 

We focus on the monocular VO problem, which takes two consecutive undistorted images $\{I_t, I_{t+1}\}$, and estimates the relative camera motion $\delta_t^{t+1} = (T, R)$, where $T \in \mathbb{R}^3 $ is the 3D translation and $R \in so(3)$ denotes the 3D rotation. 
% VO is one of the fundamental modules in a Visual SLAM system, where the output of the VO module is fed into downstream modules such as bundle adjustment and mapping. 
According to the epipolar geometry theory~\citep{nister2004efficient}, the geometry-based VO comes in two folds. Firstly, visual features are extracted and matched from $I_t$ and $I_{t+1}$. Then using the matching results, it computes the essential matrix leading to the recovery of the up-to-scale camera motion $\delta_t^{t+1}$. 

Following the same idea, our model consists of two sub-modules. One is the matching module $M_{\theta}(I_t, I_{t+1})$, estimating the dense matching result $F_t^{t+1}$ from two consecutive RGB images (i.e. optical flow). The other is a pose module $P_{\phi}(F_t^{t+1})$ that recovers the camera motion $\delta_t^{t+1}$ from the matching result (Fig.~\ref{Fig:arch}). This modular design is also widely used in other learning-based methods, especially in unsupervised VO \cite{guizilini2012semi,costante2016exploring,yin2018geonet, Zou_2018_ECCV, Ranjan_2019_CVPR}.

\subsection{Training on large scale diverse data}

The generalization capability has always been one of the most critical issues for learning-based methods. Most of the previous supervised models are trained on the KITTI dataset, which contains 11 labeled sequences and 23,201 image frames in the driving scenario~\citep{geiger2013vision}. \citet{wang2018end} presented the training and testing results on the EuRoC dataset~\citep{burri2016euroc}, collected by a micro aerial vehicle (MAV). They reported that the performance is limited by the lack of training data and the more complex dynamics of a flying robot. Surprisingly, most unsupervised methods also only train their models in very uniform scenes (e.g., KITTI and Cityscape~\citep{cordts2016cityscapes}). To our knowledge, no learning-based model has yet shown the capability of running on multiple types of scenes (car/MAV, indoor/outdoor). To achieve this, we argue that the training data has to cover diverse scenes and motion patterns. 

TartanAir~\citep{tartanair2020iros} is a large scale dataset with highly diverse scenes and motion patterns, containing more than 400,000 data frames. It provides multi-modal ground truth labels including depth, segmentation, optical flow, and camera pose. The scenes include indoor, outdoor, urban, nature, and sci-fi environments. The data is collected with a simulated pinhole camera, which moves with random and rich 6DoF motion patterns in the 3D space. 

We take advantage of the monocular image sequences $\{I_t\}$, the optical flow labels $\{F_t^{t+1}\}$, and the ground truth camera motions $\{\delta_t^{t+1}\}$ in our task. Our objective is to jointly minimize the optical flow loss $L_f$ and the camera motion loss $L_p$. The end-to-end loss is defined as: 
\begin{equation} \label{eq:loss}
L = \lambda L_f + L_p = \lambda  \|M_{\theta}(I_t, I_{t+1}) - F_t^{t+1}\| +  \|P_{\phi}(\hat{F}_t^{t+1}) - \delta_t^{t+1}\|
\end{equation}
where $\lambda$ is a hyper-parameter balancing the two losses. We use $\hat{\cdot}$ to denote the estimated variable from our model.   

Since TartanAir is purely synthetic, the biggest question is can a model learned from simulation data be generalized to real-world scenes? As discussed by \citet{tartanair2020iros}, a large number of studies show that training purely in simulation but with broad diversity, the model learned can be easily transferred to the real world. This is also known as domain randomization \cite{tobin2017domain,tremblay2018training}. In our experiments, we show that the diverse simulated data indeed enable the VO model to generalize to real-world data.

\subsection{Up-to-scale loss function}
\label{sec:loss}
The motion scale is unobservable from a monocular image sequence. In geometry-based methods, the scale is usually recovered from other sources of information ranging from known object size or camera height to extra sensors such as IMU. However, in most existing learning-based VO studies, the models generally neglect the scale problem and try to recover the motion with scale. This is feasible if the model is trained and tested with the same camera and in the same type of scenario. For example, in the KITTI dataset, the camera is mounted at a fixed height above the ground and a fixed orientation. A model can learn to remember the scale in this particular setup. Obviously, the model will have huge problems when tested with a different camera configuration. Imagine if the camera in KITTI moves a little upwards and becomes higher from the ground, the same amount of camera motion would cause a smaller optical flow value on the ground, which is inconsistent with the training data. Although the model could potentially learn to pick up other clues such as object size, it is still not fully reliable across different scenes or environments. 

Following the geometry-based methods, we only recover an up-to-scale camera motion from the monocular sequences. Knowing that the scale ambiguity only affects the translation $T$, we design a new loss function for $T$ and keep the loss for rotation $R$ unchanged. We propose two up-to-scale loss functions for $L_P$: the cosine similarity loss $L_p^{cos}$ and the normalized distance loss $L_p^{norm}$.  $L_p^{cos}$ is defined by the cosine angle between the estimated $\hat{T}$ and the label $T$: 
\begin{equation} \label{eq:sim1}
L_p^{cos} = \frac{\hat{T} \cdot T}{max( \Vert{\hat{T}} \Vert \cdot \left\Vert{T} \right\Vert, \epsilon)} + \|\hat{R} - R\|
\end{equation}

Similarly, for $L_p^{norm}$, we normalize the translation vector before calculating the distance between the estimation and the label:
\begin{equation} \label{eq:sim2}
L_p^{norm} = \left\Vert {\frac{ \hat{T}}{ max(\Vert{\hat{T}} \Vert, \epsilon)} - \frac{T}{max(\left\Vert{T} \right\Vert,\epsilon)}} \right\Vert + \|\hat{R} - R\|
\end{equation}

\noindent where $\epsilon=1\mathrm{e}\text{-}6$ is used to avoid divided by zero error. From our preliminary empirical comparison, the above two formulations have similar performance. In the following sections, we will use Eq~\ref{eq:sim2} to replace $L_p$ in Eq~\ref{eq:loss}. Later, we show by experiments that the proposed up-to-scale loss function is crucial for the model's generalization ability.

\subsection{Cross-camera generalization by encoding camera intrinsics}

In epipolar geometry theory, the camera intrinsics is required when recovering the camera pose from the essential matrix (assuming the images are undistorted). In fact, learning-based methods are unlikely to generalize to data with different camera intrinsics. Imagine a simple case that the camera changes a lens with a larger focal length. Assume the resolution of the image remains the same, the same amount of camera motion will introduce bigger optical flow values, which we call the intrinsics ambiguity. 
% \wsnote{outline what 3.4.1 and 3.4.2 talk about}

% \wsnote{assumption: images are undistorted}
A tempting solution for intrinsics ambiguity is warping the input images to match the camera intrinsics of the training data. However, this is not quite practical especially when the cameras differ too much. As shown in Fig.~\ref{Fig:intrinsics}-a, if a model is trained on TartanAir, the warped KITTI image only covers a small part of the TartanAir's field of view (FoV). After training, a model learns to exploit cues from all possible positions in the FoV and the interrelationship among those cues. Some cues no longer exist in the warped KITTI images leading to drastic performance drops. 

\subsubsection{Intrinsics layer}

\begin{figure} [t]
	\begin{center}
		\includegraphics[width=0.95\textwidth]{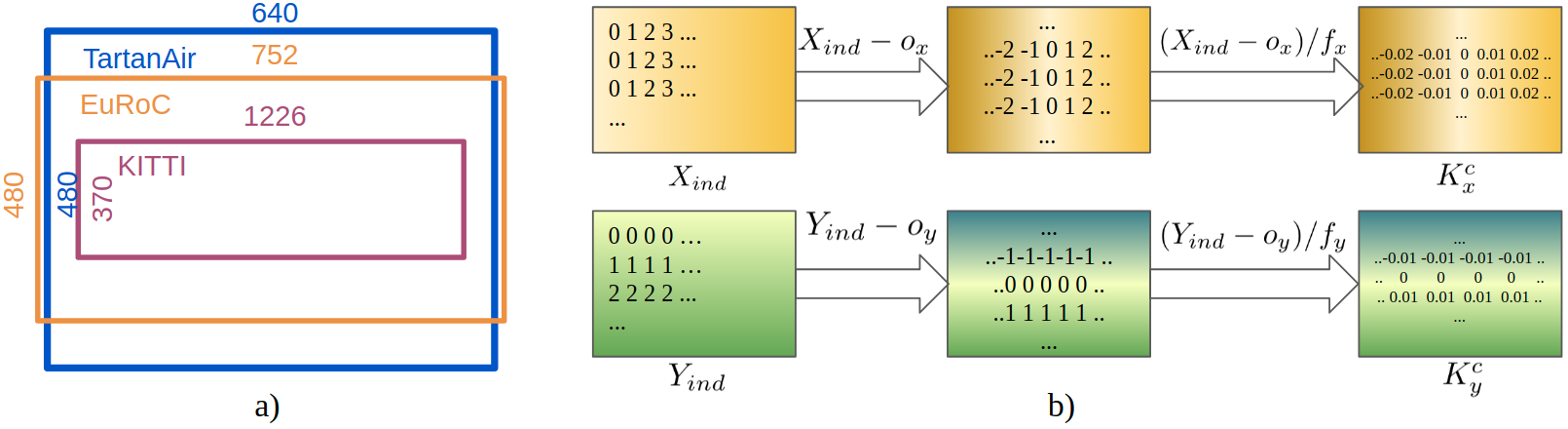}
	\end{center}
	\vspace{-0.15in}
	\caption{a) Illustration of the FoV and image resolution in TartanAir, EuRoC, and KITTI datasets. b) Calculation of the intrinsics layer. }
	\label{Fig:intrinsics}
\end{figure}

We propose to train a model that takes both RGB images and camera intrinsics as input, thus the model can directly handle images coming from various camera settings. Specifically, instead of recovering the camera motion $T_t^{t+1}$ only from the feature matching $F_t^{t+1}$, we design a new pose network $P_{\phi}(F_t^{t+1}, K) $, which depends also on the camera intrinsic parameters $K = \{ f_x, f_y, o_x, o_y \}$, 
% \begin{equation} \label{eq:K1}
% K=\begin{bmatrix} 
% f_x & 0 & o_x \\
% 0 & f_y & o_y\\
% 0 & 0 & 1 \\
% \end{bmatrix}
% \end{equation}
where $f_x$ and $f_y$ are the focal lengths, and $o_x$ and $o_y$ denote the position of the principle point. 

As for the implementation, we concatenate an IL (intrinsics layer) $K^c \in \mathbb{R}^{2 \times H \times W} $ ($H$ and $W$ are image height and width respectively) to $F_t^{t+1}$ before going into $P_{\phi}$. To compose $K^c$, we first generate two index matrices $X_{ind}$ and $Y_{ind}$ for the $x$ and $y$ axis in the 2D image frame (Fig.~\ref{Fig:intrinsics}-b). Then the two channels of $K^c$ are calculated from the following formula:  
\begin{equation} \label{eq:Intrinsic}
    \begin{cases}
    K_x^c = (X_{ind} - o_x)/f_x\\
    K_y^c = (Y_{ind} - o_y)/f_y
    \end{cases}
\end{equation}
The concatenation of $F_t^{t+1}$ and $K^c$ augments the optical flow estimation with 2D position information. Similar to the situation where geometry-based methods have to know the 2D coordinates of the matched features, $K^c$ provides the necessary position information. In this way, intrinsics ambiguity is explicitly handled by coupling 2D positions and matching estimations ($F_t^{t+1}$). 

\subsubsection{Data generation for various camera intrinsics}
\label{sec:rcr}

\begin{figure} [t]
	\begin{center}
		\includegraphics[width=0.9\textwidth]{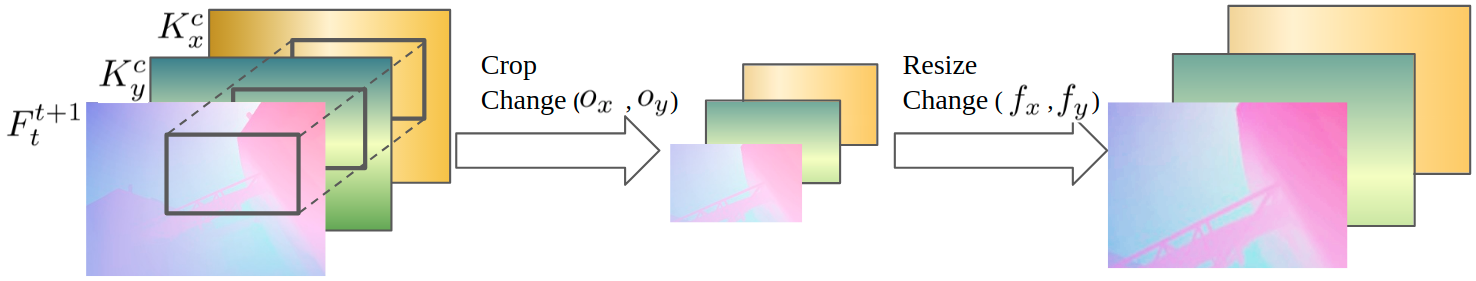}
	\end{center}
	\vspace{-0.1in}
	\caption{The data augmentation procedure of random cropping and resizing. In this way we generate a wide range of camera intrinsics (FoV $40^{\circ}$ to $90^{\circ}$).  }
	\label{Fig:rcr}
\end{figure}

To make a model generalizable across different cameras, we need training data with various camera intrinsics. TartanAir only has one set of camera intrinsics, where $f_x = f_y = 320$, $o_x=320$, and $o_y=240$. We simulate various intrinsics by randomly cropping and resizing (RCR) the input images. As shown in Fig.~\ref{Fig:rcr}, we first crop the image at a random location with a random size. Next, we resize the cropped image to the original size. One advantage of the IL is that during RCR, we can crop and resize the IL with the image, without recomputing the IL. To cover typical cameras with FoV between $40^{\circ}$ to $90 ^{\circ}$, we find that using random resizing factors up to 2.5 is sufficient during RCR. Note the ground truth optical flow should also be scaled with respect to the resizing factor. We use very aggressive cropping and shifting in our training, which means the optical center could be way off the image center. Although the resulting intrinsic parameters will be uncommon in modern cameras, we find the generalization is improved.

%===============================================================================

\section{Experimental Results}
\label{sec:result}
% In this section, we first introduce the implementation details in Sec~\ref{sec:impl} followed by experimental analysis of the three main contributions. In Sec~\ref{sec:gen}, we evaluate the generalization ability of the proposed model by testing it on various datasets \textit{without any finetuing on the target dataset}. We compare our model with various state-of-the-art algorithms on multiple public datasets, and challenging testing trajectories provided by TartanAir. 

% Next, we analyze the effectiveness of three main contributions of this paper, which are the large and diverse training data, the up-to-scale loss function and the novel IL in Sec~\ref{sec:data} - \ref{sec:intrin}.  

% In Sec~\ref{sec:gen}, we evaluate the generalization ability of the proposed model by testing it on various datasets \textit{without any finetuing on the target dataset}. We compare our model with various state-of-the-art algorithms in different scenarios, including outdoor driving scenario from KITTI, indoor MAV data from EuRoC, sequences collected by ourselves, and challenging testing trajectories provided by TartanAir. 
 
\subsection{Network structure and training detail}
\label{sec:impl}
\paragraph{Network} We utilize the pre-trained PWC-Net~\citep{sun2018pwc} as the matching network $M_{\theta}$, and a modified ResNet50~\citep{he2016deep} as the pose network $P_{\phi}$.  We remove the batch normalization layers from the ResNet, and add two output heads for the translation and rotation, respectively. The PWC-Net outputs optical flow in size of $H/4 \times W/4$, so $P_{\phi}$ is trained on 1/4 size, consuming very little GPU memory. The overall inference time (including both $M_{\theta}$ and $P_{\phi}$) is 40ms on an NVIDIA GTX 1080 GPU.
\vspace{-0.1in}
\paragraph{Training} Our model is implemented by PyTorch~\citep{paszke2017automatic} and trained on 4 NVIDIA GTX 1080 GPUs. There are two training stages. First, $P_{\phi}$ is trained separately using ground truth optical flow and camera motions for 100,000 iterations with a batch size of 100. In the second stage, $P_{\phi}$ and $M_{\theta}$ are connected and jointly optimized for 50,000 iterations with a batch size of 64. During both training stages, the learning rate is set to $1\mathrm{e}\text{-}4$ with a decay rate of 0.2 at 1/2 and 7/8 of the total training steps. The RCR is applied on the optical flow, RGB images, and the IL (Sec~\ref{sec:rcr}). 

\subsection{How the training data quantity affects the generalization ability}
\label{sec:data}
% Generalization is a critical problem for applying the learning-based methods to real applications. 

\begin{figure} [t]
	\begin{center}
		\includegraphics[width=1.0\textwidth]{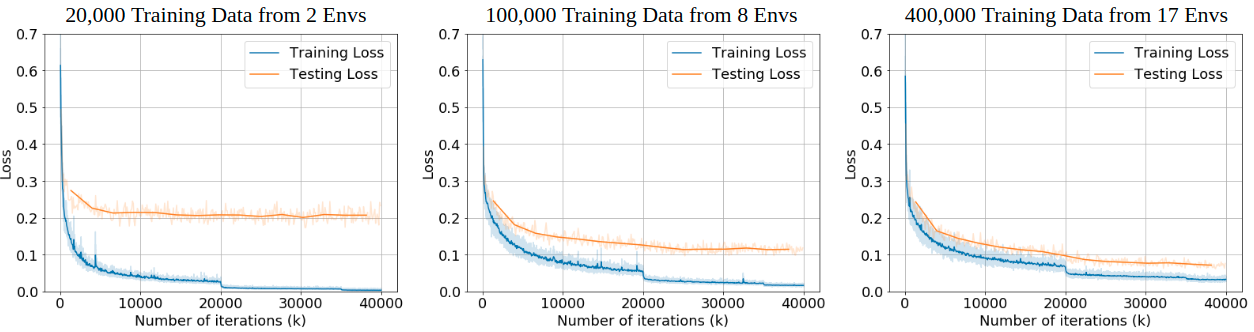}
	\end{center}
	\vspace{-0.1in}
	\caption{Generalization ability with respect to different quantities of training data. Model $P_{\phi}$ is trained on true optical flow. Blue: training loss, orange: testing loss on three unseen environments. Testing loss drops constantly with increasing quantity of training data.}
	\label{Fig:diverse}
\end{figure}

To show the effects of data diversity, we compare the generalization ability of the model trained with different amounts of data. We use 20 environments from the TartanAir dataset, and set aside 3 environments (Seaside-town, Soul-city, and Hongkong-alley) only for testing, which results in more than 400,000 training frames and about 40,000 testing frames. As a comparison, KITTI and EuRoC datasets provide 23,201 and 26,604 pose labeled frames, respectively. Besides, data in KITTI and EuRoC are much more uniform in the sense of scene type and motion pattern. As shown in Fig.~\ref{Fig:diverse}, we set up three experiments, where we use 20,000 (comparable to KITTI and EuRoC), 100,000, and 400,000 frames of data for training the pose network $P_{\phi}$. The experiments show that the generalization ability, measured by the gap between training loss and testing loss on unseen environments, improves constantly with increasing training data. 

\subsection{Up-to-scale loss function}
\begin{figure} [t]
	\begin{center}
		\includegraphics[width=1.0\textwidth]{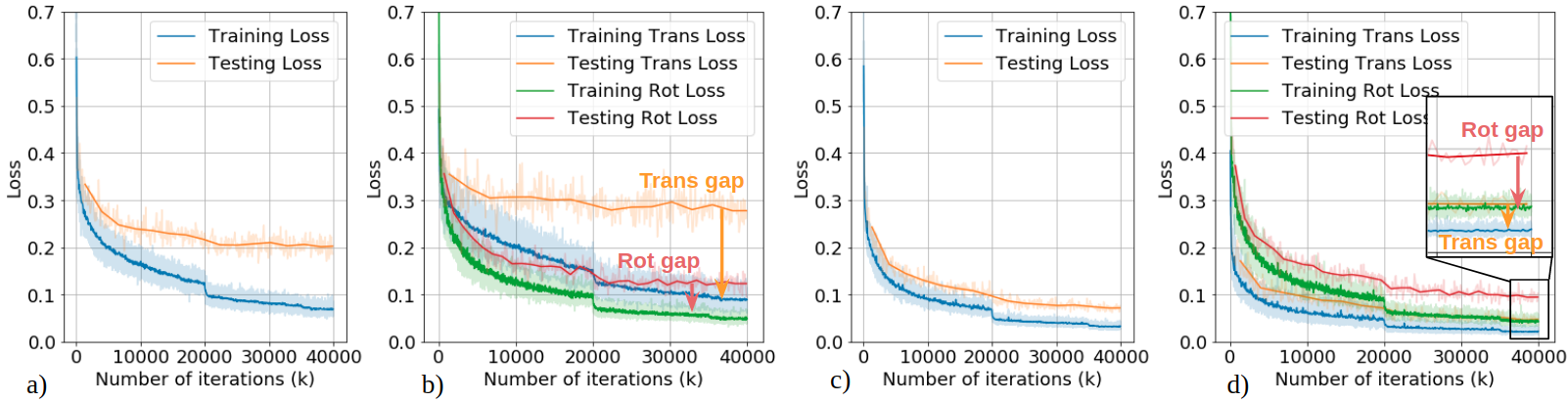}
	\end{center}
	\vspace{-0.1in}
	\caption{Comparison of the loss curve w/ and w/o up-to-scale loss function. a) The training and testing loss w/o the up-to-scale loss. b) The translation and rotation loss of a). Big gap exists between the training and testing translation losses (orange arrow in b)). c) The training and testing losses w/ up-to-scale loss. d) The translation and rotation losses of c). The translation loss gap decreases.}
	\vspace{-0.1in}
	\label{Fig:scale}
\end{figure}

Without the up-to-scale loss, we observe that there is a gap between the training and testing loss even trained with a large amount of data (Fig.~\ref{Fig:scale}-a). As we plotting the translation loss and rotation loss separately (Fig.~\ref{Fig:scale}-b), it shows that the translation error is the main contributor to the gap. After we apply the up-to-scale loss function described in Sec~\ref{sec:loss}, the translation loss gap decreases (Fig.~\ref{Fig:scale}-c,d). During testing, we align the translation with the ground truth to recover the scale using the same way as described in \cite{yin2018geonet, Zhou2017SfMLearner}. 

% \wsnote{testing traj w/ and w/o up-to-scale loss}

\subsection{Camera intrinsics layer}
\label{sec:intrin}
% \subsubsection{Robustness with inaccurate camera intrinsics}

The IL is critical to the generalization ability across datasets. Before we move to other datasets, we first design an experiment to investigate the properties of the IL using the pose network $P_{\phi}$. As shown in Table~\ref{tab:intrinsics}, in the first two columns, where the data has no RCR augmentation, the training and testing loss are low. But these two models would output nonsense values on data with RCR augmentation. One interesting finding is that adding IL doesn't help in the case of only one type of intrinsics. This indicates that the network has learned a very different algorithm compared with the geometry-based methods, where the intrinsics is necessary to recover the motion. The last two columns show that the IL is critical when the input data is augmented by RCR (i.e. various intrinsics). Another interesting thing is that training a model with RCR and IL leads to a lower testing loss (last column) than only training on one type of intrinsics (first two columns). This indicates that by generating data with various intrinsics, we learned a more robust model for the VO task. 

\begin{table}[t!]
\small
\caption{Training and testing losses with four combinations of RCR and IL settings. The IL is critical with the presence of RCR. The model trained with RCR reaches lower testing loss than those without RCR. }
\begin{center}
\begin{tabular}{lcccc}
\hline
Training configuration      & w/o RCR, w/o IL & w/o RCR,  w/ IL  & w/ RCR, w/o IL & w/ RCR, w/ IL \\ \hline
Training loss               & 0.0325 & 0.0311 & 0.1534     & 0.0499    \\ \hline
Test-loss on data w/ RCR  &  -     & -      & 0.1999     & 0.0723    \\
Test-loss on data w/o RCR & 0.0744 & 0.0714 & 0.1630     & 0.0549    \\ \hline
\end{tabular}
\end{center}
\label{tab:intrinsics}
\end{table}

% \wsnote{move tartan testing here}
% \wsnote{crop and resize differently and plot the trajectories}

% \wsnote{the variance reflect the uncertainty}

% \begin{table}[]
% \small
% \caption{Placeholder, Comparison of translation and rotation errors between our method and other visual odometry methods on the KITTI benchmark.}
% \begin{center}
% \begin{tabular}{lcccc}
% \hline
%                      & \multicolumn{1}{l}{w/o IL} & \multicolumn{1}{l}{w/ IL} & \multicolumn{1}{l}{RCR w/o IL} & \multicolumn{1}{l}{RCR w/ IL} \\ \hline
% Training Loss        & 0.0325                     & 0.0311                    & 0.1534                         & 0.0499                        \\
% Testing Loss         & 0.0744                     & 0.0714                    & 0.1999                         & 0.0723                        \\
% Testing Loss w/o RCR & -                          & -                         & 0.163                          & 0.0549                        \\ \hline
% \end{tabular}
% \end{center}
% \label{tab:intrinsics}
% \end{table}

\subsection{Generalize to real-world data without finetuning}
\label{sec:gen}

\begin{table*}[]
\vspace{-0.1in}
\small
\caption{Comparison of translation and rotation on the KITTI dataset. DeepVO~\citep{wang2017deepvo} is a supervised method trained on Seq 00, 02, 08, 09. It contains an RNN module, which accumulates information from multiple frames. \citet{wang2019improving} is a supervised method trained on Seq 00-08 and uses the semantic information of multiple frames to optimize the trajectory. UnDeepVO~\citep{li2018undeepvo} and GeoNet~\citep{yin2018geonet} are trained on Seq 00-08 in an unsupervised manner. VISO2-M~\citep{Song2015MoncularScale} and ORB-SLAM~\citep{mur2015orb} are geometry-based monocular VO. ORB-SLAM uses the bundle adjustment on multiple frames to optimize the trajectory. Our method works in a pure VO manner (only takes two frames). It has never seen any KITTI data before the testing, and yet achieves competitive results. }
\vspace{-0.1in}
\begin{center}
\small
\begin{tabular}{lcccccccccc}
\hline
\multicolumn{1}{c}{Seq} & \multicolumn{2}{c}{06}         & \multicolumn{2}{c}{07}         & \multicolumn{2}{c}{09}         & \multicolumn{2}{c}{10}       & \multicolumn{2}{c}{Ave}      \\            & $t_{rel}$                          & \multicolumn{1}{c}{$r_{rel}$} & $t_{rel}$                  & \multicolumn{1}{c}{$r_{rel}$}       & $t_{rel}$                          & \multicolumn{1}{c}{$r_{rel}$} & $t_{rel}$                          & \multicolumn{1}{c}{$r_{rel}$} & $t_{rel}$                          & \multicolumn{1}{c}{$r_{rel}$} \\ \hline
DeepVO~\citep{wang2017deepvo}*$\dagger$           & 5.42          & 5.82          & 3.91          & 4.6           & -             & -             & 8.11          & 8.83         & 5.81          & 6.41         \\
Wang et al.~\citep{wang2019improving}*$\dagger$      & -             & -             & -             & -             & 8.04          & 1.51          & 6.23          & 0.97         & 7.14         & 1.24         \\
UnDeepVO~\citep{li2018undeepvo}*              & 6.20          & 1.98          & \textbf{3.15} &  2.48 & -             & -             & 10.63         & 4.65         & 6.66          & 3.04         \\
GeoNet~\citep{yin2018geonet}*                 & 9.28          & 4.34          & 8.27          & 5.93          & 26.93         & 9.54          & 20.73         & 9.04         & 16.3          & 7.21         \\
VISO2-M~\citep{Song2015MoncularScale}             & 7.3           & 6.14          & 23.61         & 19.11         & \textbf{4.04}          & 1.43          & 25.2          & 3.8          & 15.04         & 7.62         \\
ORB-SLAM~\citep{mur2015orb}$\dagger$       & 18.68         & \textbf{0.26} & 10.96         & \textbf{0.37} & 15.3          & \textbf{0.26} & \textbf{3.71} & \textbf{0.3} & 12.16         & \textbf{0.3} \\
TartanVO (ours)                & \textbf{4.72} & 2.95          & 4.32          & 3.41          & 6.0    & 3.11           & 6.89          & 2.73         & \textbf{5.48} & 3.05         \\ \hline
\end{tabular}
\end{center}
\begin{tablenotes}
      \small
      \item $t_{rel}$: average translational RMSE drift (\%) on a length of 100–800 m.
      \item $r_{rel}$: average rotational RMSE drift ($^\circ$/100 m) on a length of 100–800 m.
      \item *: the starred methods are trained or finetuned on the KITTI dataset.
      \item $\dagger$: these methods use multiple frames to optimize the trajectory after the VO process.
     \end{tablenotes}
	 \vspace{-0.1in}
\label{tab:kitti_compare}
\end{table*}

\paragraph{KITTI dataset} The KITTI dataset is one of the most influential datasets for VO/SLAM tasks.
% , consists of 11 pose-labeled sequences in ourdoor driving scenario. 
We compare our model, TartanVO, with two supervised learning models (DeepVO~\citep{wang2017deepvo}, \citet{wang2019improving}), two unsupervised models (UnDeepVO~\citep{li2018undeepvo}, GeoNet~\citep{yin2018geonet}), and two geometry-based methods (VISO2-M~\citep{Song2015MoncularScale}, ORB-SLAM~\citep{mur2015orb}).
% Most supervised learning methods train on part of the 11 labeled sequences and evaluate on the rest. 
All the learning-based methods except ours are trained on the KITTI dataset. 
Note that our model has not been finetuned on KITTI and is trained purely on a synthetic dataset. 
Besides, many algorithms use multiple frames to further optimize the trajectory. In contrast, our model only takes two consecutive images. As listed in Table~\ref{tab:kitti_compare}, TartanVO achieves comparable performance, despite no finetuning nor backend optimization are performed.

\begin{table}[t]
\small
\caption{Comparison of ATE on EuRoC dataset. We are among very few learning-based methods, which can be tested on this dataset. Same as the geometry-based methods, our model has never seen the EuRoC data before testing. We show the best performance on two difficult sequences VR1-03 and VR2-03. Note our method doesn't contain any backend optimization module.}
\begin{center}
\small
\begin{tabular}{llcccccc}
\hline
                                & Seq.            & MH-04 & MH-05 & VR1-02 & VR1-03 & VR2-02 & VR2-03 \\ \hline
\multirow{4}{*}{Geometry-based *} & SVO~\citep{forster2016svo}             & 1.36  & 0.51  & 0.47   & x      & 0.47   & x      \\
                                & ORB-SLAM~\citep{mur2015orb}        & \textbf{0.20}  & 0.19  & x      & x      & \textbf{0.07}   & x      \\
                                & DSO~\citep{engel2017direct}             & 0.25  & \textbf{0.11}  & \textbf{0.11}   & 0.93   & 0.13   & 1.16   \\
                                & LSD-SLAM~\citep{Engel2014lsd}        & 2.13  & 0.85  & 1.11   & x      & x      & x      \\ \hline
Learning-based $\dagger$                  & TartanVO (ours) & 0.74  & 0.68  & 0.45   & \textbf{0.64}   & 0.67   & \textbf{1.04}   \\ \hline
\end{tabular}
\end{center}
\vspace{-0.1in}
\begin{tablenotes}
      \small
      \item \quad * These results are from \citep{forster2016svo}. $\dagger$ Other learning-based methods \citep{burri2016euroc} did not report numerical results.
\end{tablenotes}
\label{tab:euroc}     
\end{table}

\begin{figure} [t]
	\begin{center}
		\includegraphics[width=1.0\textwidth]{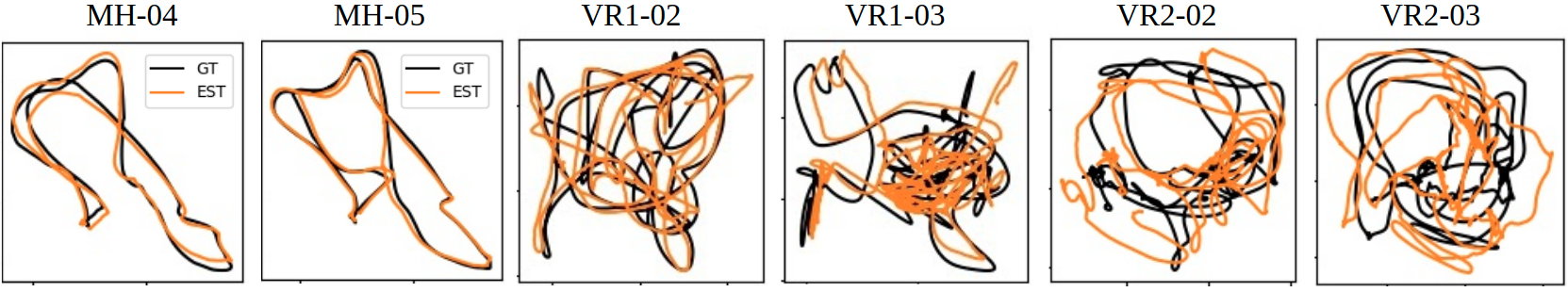}
	\end{center}
	\vspace{-0.1in}
	\caption{The visualization of 6 EuRoC trajectories in Table~\ref{tab:euroc}. Black: ground truth trajectory, orange: estimated trajectory. }
	\label{Fig:euroc}
	\vspace{-0.1in}
\end{figure}

\vspace{-0.1in}
\paragraph{EuRoC dataset} The EuRoC dataset contains 11 sequences collected by a MAV in an indoor environment. There are 3 levels of difficulties with respect to the motion pattern and the light condition. Few learning-based methods have ever been tested on EuRoC due to the lack of training data. 
The changing light condition and aggressive rotation also pose real challenges to geometry-based methods as well. 
In Table~\ref{tab:euroc}, we compare with geometry-based methods including SVO~\citep{forster2016svo}, ORB-SLAM~\citep{mur2015orb}, DSO~\citep{engel2017direct} and LSD-SLAM~\citep{Engel2014lsd}. Note that all these geometry-based methods perform some types of backend optimization on selected keyframes along the trajectory. 
In contrast, our model only estimates the frame-by-frame camera motion, which could be considered as the frontend module in these geometry-based methods. In Table~\ref{tab:euroc}, we show the absolute trajectory error (ATE) of 6 medium and difficult trajectories. Our method shows the best performance on the two most difficult trajectories VR1-03 and VR2-03, where the MAV has very aggressive motion. A visualization of the trajectories is shown in Fig.~\ref{Fig:euroc}.

\begin{table}[h!]
\caption{Comparison of ATE on TartanAir dataset. These trajectories are not contained in the training set. We repeatedly run ORB-SLAM 5 times and report the best result. }
\small
\begin{tabular}{lcccccccc}
\hline
Seq      & MH000        & MH001         & MH002      & MH003         & MH004         & MH005         & MH006      & MH007         \\\hline
ORB-SLAM~\citep{mur2015orb} & \textbf{1.3} & \textbf{0.04} & 2.37       & 2.45          & x             & x             & 21.47      & 2.73          \\
TartanVO (ours) & 4.88         & 0.26          & \textbf{2} & \textbf{0.94} & \textbf{1.07} & \textbf{3.19} & \textbf{1} & \textbf{2.04} \\
\hline
\end{tabular}
\label{tab:tartanair}
\end{table}

% \begin{table}[]
% \caption{Comparison of ATE error on EuRoC dataset. }
% \begin{center}
% \small
% \begin{tabular}{lcccccc}
% \hline
% Seq     & MH-04 & MH-05 & VR1-02 & VR1-03 & VR2-02 & VR2-03 \\ \hline
% SVO~\citep{forster2016svo}      & 1.36  & 0.51  & 0.47   & x      & 0.47   & x      \\
% ORB-SLAM~\citep{mur2015orb} & 0.20  & 0.19  & x      & x      & 0.07   & x      \\
% DSO~\citep{engel2017direct}      & 0.25  & 0.11  & 0.11   & 0.93   & 0.13   & 1.16   \\
% LSD-SLAM~\citep{Engel2014lsd} & 2.13  & 0.85  & 1.11   & x      & x      & x      \\ 
% TartanVO(ours)  & 0.74  & 0.51  & 0.44   & 0.8098 & 1.0429 & 1.4336 \\ \hline
% \end{tabular}
% \end{center}
% \vspace{-0.1in}
% \begin{tablenotes}
%       \small
%       \item \qquad\qquad \qquad the results of geometry-based methods are from \citep{forster2016svo}
%      \end{tablenotes}

% \end{table}

% \begin{figure} [t]
% 	\begin{center}
% 		\includegraphics[width=1.0\textwidth]{figure/euroc_results.png}
% 	\end{center}
% 	\caption{Placeholder, translation loss }
% 	\label{Fig:euroc}
% \end{figure}

\vspace{-0.1in}
\paragraph{Challenging TartanAir data} 
TartanAir provides 16 very challenging testing trajectories\footnote{https://github.com/castacks/tartanair\_tools\#download-the-testing-data-for-the-cvpr-visual-slam-challenge} that cover many extremely difficult cases, including changing illumination, dynamic objects, fog and rain effects, lack of features, and large motion. As listed in Table~\ref{tab:tartanair}, we compare our model with the ORB-SLAM using ATE. Our model shows a more robust performance in these challenging cases. 

\begin{figure}[t!]
    \centering
    \includegraphics[width=1.\textwidth]{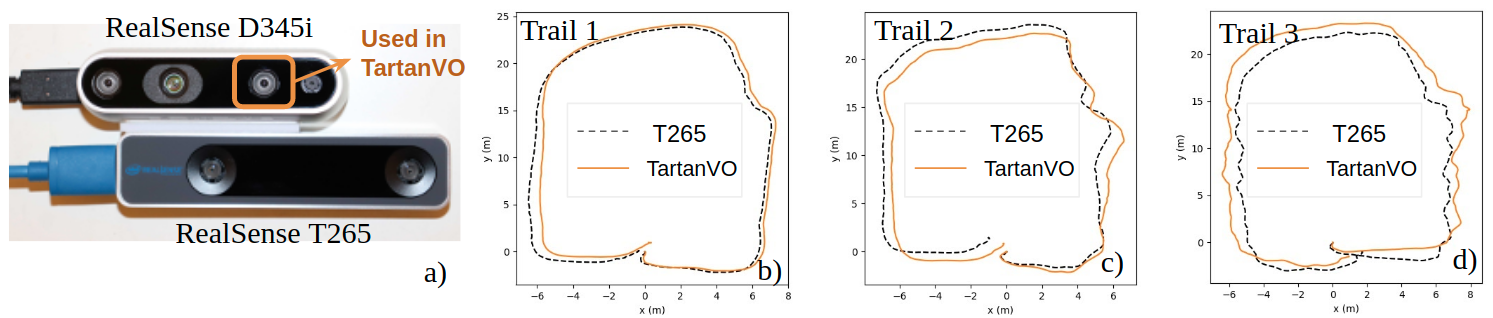}
    \vspace{-0.2in}
    \caption{
    % Visualization of 3 trajectories collected by a customized sensory setup. Those trajectories are loop trails with increasing difficulty in the motion pattern. The black dashed line shows the estimated camera motion output by the RealSense T265 tracking camera. The orange line shows the estimated trajectory by TartanVO. 
    TartanVO outputs competitive results on D345i IR data compared to T265 (equipped with fish-eye stereo camera and an IMU). a) The hardware setup. b) Trail 1: smooth and slow motion. c) Trail 2: smooth and medium speed. d) Trail 3: aggressive and fast motion. See videos for details.
    }
    \label{fig:realsense}
    \vspace{-0.15in}
\end{figure}

\vspace{-0.1in}
\paragraph{RealSense Data Comparison}
We test TartanVO using data collected by a customized sensor setup. 
As shown in Fig.~\ref{fig:realsense}~a), a RealSense D345i is fixed on top of a RealSense T265 tracking camera. We use the left near-infrared (IR) image on D345i in our model and compare it with the trajectories provided by the T265 tracking camera.   
We present 3 loopy trajectories following similar paths with increasing motion difficulties. From Fig.~\ref{fig:realsense}~b) to d), we observe that although TartanVO has never seen real-world images or IR data during training, it still generalizes well and predicts odometry closely matching the output of T265, which is a dedicated device estimating the camera motion with a pair of fish-eye stereo camera and an IMU. 

\vspace{-0.1in}
% \paragraph{Realsense Data}

% Since AlphaVO is a VO, it does not
% detect loop closures and only maintains a small local map of
% the last five to ten keyframes. Additionally, we provide results
% with the same configuration using both image streams of the
% stereo camera. Therefore, we apply the approach introduced in
% Section VIII to estimate the motion of a multicamera system.

% https://ieeexplore.ieee.org/stamp/stamp.jsp?arnumber=7782863&casa_token=pOhtjMLck_sAAAAA:nXdBf050epszLxCq6foPoPRHtmsSFlDUH80YrkpH5hBcx9Uf2JUGbYlw8EjXPWFuzovYacHVCg
% https://ieeexplore.ieee.org/stamp/stamp.jsp?arnumber=8421746&casa_token=0fVD4E7a8RIAAAAA:jnd8VXww7Aln6Hy-y9V0lmOzcc4YsKionA-dSN7EeZ6NkSWqxjL8UoBNj-njM36DBVJynMHcUg
%https://journals.sagepub.com/doi/pdf/10.1177/0278364917734298?casa_token=TnD2VGIk5bYAAAAA:QtLxH3MykvGLtNb7_3NinduQapf_eA4tn2JXInkkZJq6GhWvu6-qQB77dF-1bvYjLTB_XGFRucgasA

% GT flow w/ and w/o ILs
% check

% baselines: orb-mono, orb-stero, viso, etc (copy from other's paper)

% EuRoC

% baselines: copy from other's paper

% \subsection{Lesson learned}

%===============================================================================

\section{Conclusions}
\label{sec:conclusion}
\vspace{-0.1in}

We presented TartanVO, a generalizable learning-based visual odometry. By training our model with a large amount of data, we show the effectiveness of diverse data on the ability of model generalization. A smaller gap between training and testing losses can be expected with the newly defined up-to-scale loss, further increasing the generalization capability. We show by extensive experiments that, equipped with the intrinsics layer designed explicitly for handling different cameras, TartanVO can generalize to unseen datasets and achieve performance even better than dedicated learning models trained directly on those datasets. Our work introduces many exciting future research directions such as generalizable learning-based VIO, Stereo-VO, multi-frame VO.

%===============================================================================

% The maximum paper length is 8 pages excluding references and acknowledgements, and 10 pages including references and acknowledgements

\clearpage
% The acknowledgments are automatically included only in the final version of the paper.
\acknowledgments{This work was supported by ARL award \#W911NF1820218. Special thanks to Yuheng Qiu and Huai Yu from Carnegie Mellon University for preparing simulation results and experimental setups. }

%===============================================================================

% no \bibliographystyle is required, since the corl style is automatically used.% 
\small
% \bibliography{}  % .bib
\setlength{\bibsep}{4pt plus 0.3ex}
{\linespread{0.7}\selectfont\bibliography{example}}
% \bibliography{example}
\normalsize

\appendix
\section{Additional experimental details}
In this section, we provide additional details for the experiments, including the network structure, training parameters, qualitative results, and quantitative results.

\subsection{Network Structure}
Our network consists of two sub-modules, namely, the matching network $M_{\theta}$ and the pose network $P_{\phi}$. As mentioned in the paper, we employ PWC-Net as the matching network, which takes in two consecutive images of size 640 x 448 (PWC-Net only accepts image size that is multiple of 64). The output optical flow, which is 160 x 112 in size, is fed into the pose network. The structure of the pose network is detailed in Table~\ref{tab:network}. The overall inference time (including both $M_{\theta}$ and $P_{\phi}$) is 40ms on an NVIDIA GTX 1080 GPU. 

\begin{table}[!hb]
\caption{Parameters of the proposed pose network. Constructions of residual blocks are designated in brackets multiplied by the number of stacked blocks. Downsampling is performed by Conv1, and at the beginning of each residual block. After the residual blocks, we reshape the feature map into a one-dimensional vector, which goes through three fully connected layers in the translation head and rotation head, respectively. }
\begin{tabular}{|M{2.5cm}|M{2.7cm}|M{3cm}|M{2.7cm}|N}
\hline
Name                         & Layer setting                            & \multicolumn{2}{c|}{Output dimension}                                     &  \\ \hline
Input                        &                                          & $\frac{1}{4}H \times \frac{1}{4}W \times 2$       & $114 \times 160$      &  \\ [2pt] \hline
Conv1 & $3 \times 3 \text{, } 32$                & $\frac{1}{8}H \times \frac{1}{8}W \times 32$      & $56 \times 80$        &  \\[2pt] \hline
Conv2 & $3 \times 3 \text{, } 32$                & $\frac{1}{8}H \times \frac{1}{8}W \times 32$      & $56 \times 80$        &  \\[2pt] \hline
Conv3 & $3 \times 3 \text{, } 32$                & $\frac{1}{8}H \times \frac{1}{8}W \times 32$      & $56 \times 80$        &  \\[2pt] \hline
\multicolumn{4}{|c|}{ResBlock}                       &  \\ \hline
Block1                       & $\begin{bmatrix} 3 \times 3 \text{, } 64 \\ 3 \times 3 \text{, } 64 \end{bmatrix} \times 3$    & $\frac{1}{16}H \times \frac{1}{16}W \times 64$    & $ 28 \times 40 $      &  \\ \hline
Block2                       & $\begin{bmatrix} 3 \times 3 \text{, } 128 \\ 3 \times 3 \text{, } 128 \end{bmatrix} \times 4$    & $\frac{1}{32}H \times \frac{1}{32}W \times 128$   & $ 14 \times 20 $      &  \\ \hline
Block3                       & $\begin{bmatrix} 3 \times 3 \text{, } 128 \\ 3 \times 3 \text{, } 128 \end{bmatrix} \times 6$    & $\frac{1}{64}H \times \frac{1}{64}W \times 128$   & $ 7 \times 10 $       &  \\ \hline
Block4                       & $\begin{bmatrix} 3 \times 3 \text{, } 256 \\ 3 \times 3 \text{, } 256 \end{bmatrix} \times 7$    & $\frac{1}{128}H \times \frac{1}{128}W \times 256$ & $ 4 \times 5 $        &  \\ \hline
Block5                       & $\begin{bmatrix} 3 \times 3 \text{, } 256 \\ 3 \times 3 \text{, } 256 \end{bmatrix} \times 3$    & $\frac{1}{256}H \times \frac{1}{256}W \times 256$ & $ 2 \times 3 $        &  \\ \hline
\multicolumn{2}{|c|}{FC\_trans}                                         & \multicolumn{2}{c|}{FC\_rot}                                              &  \\ \hline
Trans\_head\_fc1 & $\langle 256 \times 6 \rangle \times 128$ & Rot\_head\_fc1 & $ \langle 256 \times 6 \rangle \times 128 $                        &  \\ \hline
Trans\_head\_fc2 & $128 \times 32$                          & Rot\_head\_fc2 & $128 \times 32$                       &  \\ \hline
Trans\_head\_fc3 & $ 32 \times 3 $                          & Rot\_head\_fc3 & $32 \times 3$ &  \\ \hline
Output & $ 3 $ & Output & $3$ &  \\ \hline
\end{tabular}
\label{tab:network}
\end{table}

% \subsection{Training Parameters}

\begin{table}[h]
\caption{Comparison of ORB-SLAM and TartanVO on the TartanAir dataset using the ATE metric. These trajectories are not contained in the training set. We repeatedly run ORB-SLAM for 5 times and report the best result. }
\small
\begin{tabular}{lcccccccc}
\hline
Seq      & SH000        & SH001         & SH002      & SH003         & SH004         & SH005         & SH006      & SH007         \\\hline
ORB-SLAM & x     & 3.5     & x       & x          & x             & x             & x      & x          \\
TartanVO (ours) & \textbf{2.52}   & \textbf{1.61}    & \textbf{3.65} & \textbf{0.29} & \textbf{3.36} & \textbf{4.74} & \textbf{3.72} & \textbf{3.06} \\
\hline
\end{tabular}
\label{tab:tartanair2}
\end{table}

\begin{figure}[!hb]
    \centering
    \includegraphics[width=1.0\textwidth]{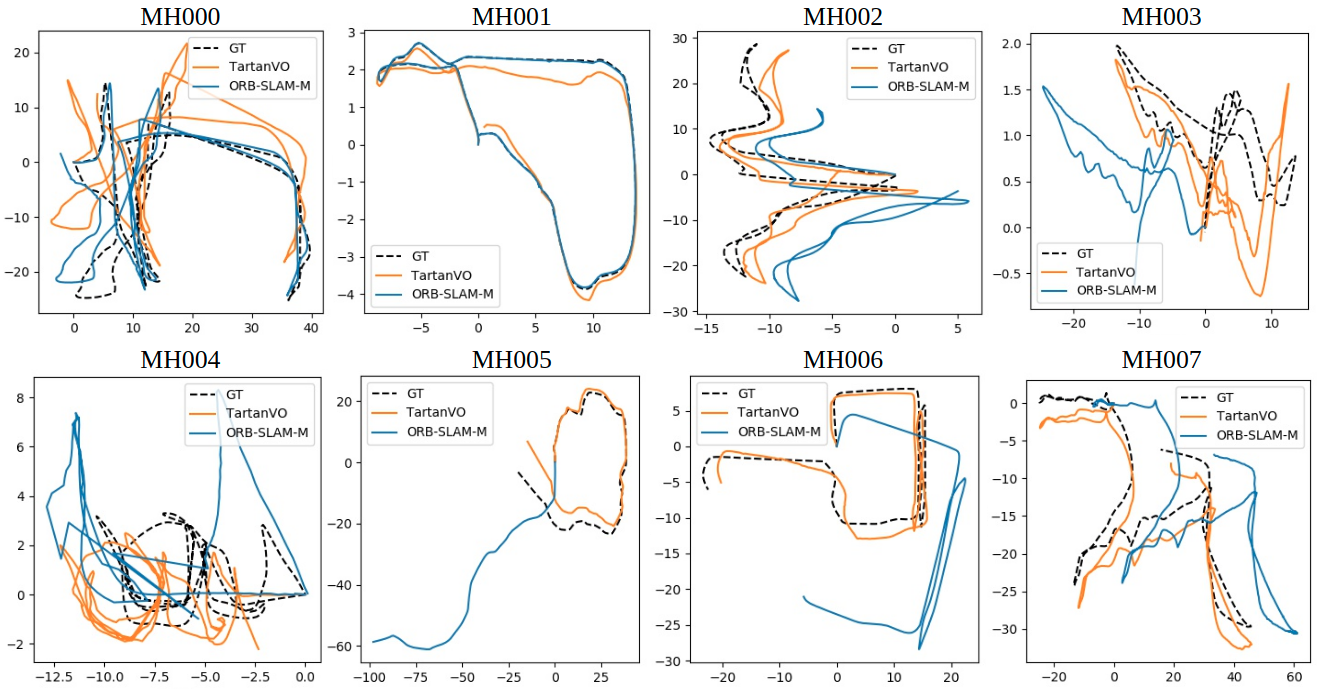}
    \includegraphics[width=1.0\textwidth]{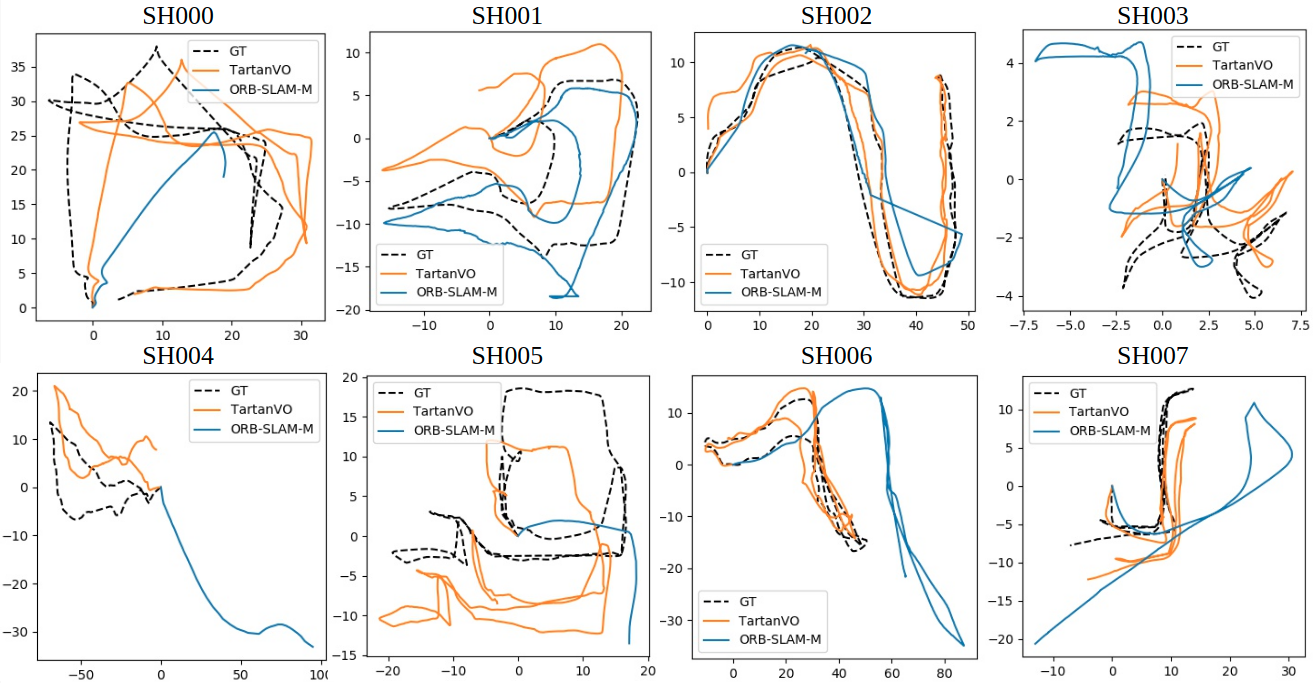}
    \caption{Visualization of the 16 testing trajectories in the TartanAir dataset. The black dashed line represents the ground truth. The estimated trajectories by TartanVO and the ORB-SLAM monocular algorithm are shown in orange and blue lines, respectively. The ORB-SLAM algorithm frequently loses tracking in these challenging cases. It fails in 9/16 testing trajectories. Note that we run full-fledge ORB-SLAM with local bundle adjustment, global bundle adjustment, and loop closure components. In contrast, although TartanVO only takes in two images, it is much more robust than ORB-SLAM.  }
    \label{fig:tartanair}
\end{figure}

\subsection{Testing Results on TartanAir}

TartanAir provides 16 challenging testing trajectories. We reported 8 trajectories in the experiment section. The rest 8 trajectories are shown in Table~\ref{tab:tartanair2}. We compare TartanVO against the ORB-SLAM monocular algorithm. Due to the randomness in ORB-SLAM, we repeatedly run ORB-SLAM for 5 trials and report the best result. We consider a trial is a failure if ORB-SLAM tracks less than 80\% of the trajectory. A visualization of all the 16 trajectories (including the 8 trajectories shown in the experiment section) is shown in Figure~\ref{fig:tartanair}.

\end{document}